\documentclass[runningheads]{llncs}
\usepackage{graphicx}
\usepackage{authblk}
\usepackage{graphicx}
\usepackage{subfigure}
\usepackage{array}
\usepackage{float}
\usepackage{booktabs}
\usepackage{bbding}
\usepackage{etoolbox}
\usepackage{color}
\usepackage{soul}
\usepackage{algorithm}
\usepackage{algorithmic}
\usepackage{cite}
\usepackage{balance}
\usepackage{amsmath, amssymb}
\usepackage{makecell}
\usepackage{multirow}
\usepackage{hyperref}

\begin{document}
\title{CAT-ViL: Co-Attention Gated Vision-Language Embedding for Visual Question Localized-Answering in Robotic Surgery}
\titlerunning{CAT-ViL for Visual Question Localized-Answering in Robotic Surgery}

\authorrunning{L. Bai et al.}

\author{Long Bai\inst{1~\star} 
\and Mobarakol Islam\inst{2} 
\thanks{Long Bai and Mobarakol Islam are co-first authors.} 
\and Hongliang Ren\inst{1,3} 
\thanks{Corresponding author.}}
\institute{Department of Electronic Engineering, The Chinese University of Hong Kong (CUHK), Hong Kong SAR, China
\and Wellcome/EPSRC Centre for Interventional and Surgical Sciences (WEISS), University College London, London, UK
\and Shun Hing Institute of Advanced Engineering, CUHK, Hong Kong SAR, China\\ 
\email{b.long@link.cuhk.edu.hk, mobarakol.islam@ucl.ac.uk, hlren@ee.cuhk.edu.hk}}

\maketitle             
\begin{abstract}

Medical students and junior surgeons often rely on senior surgeons and specialists to answer their questions when learning surgery. However, experts are often busy with clinical and academic work, and have little time to give guidance. Meanwhile, existing deep learning (DL)-based surgical Visual Question Answering (VQA) systems can only provide simple answers without the location of the answers. In addition, vision-language (ViL) embedding is still a less explored research in these kinds of tasks. Therefore, a surgical Visual Question Localized-Answering (VQLA) system would be helpful for medical students and junior surgeons to learn and understand from recorded surgical videos. We propose an end-to-end Transformer with the \textbf{C}o-\textbf{A}ttention ga\textbf{T}ed \textbf{Vi}sion-\textbf{L}anguage (CAT-ViL) embedding for VQLA in surgical scenarios, which does not require feature extraction through detection models. The CAT-ViL embedding module is designed to fuse multimodal features from visual and textual sources. The fused embedding will feed a standard Data-Efficient Image Transformer (DeiT) module, before the parallel classifier and detector for joint prediction. We conduct the experimental validation on public surgical videos from MICCAI EndoVis Challenge 2017 and 2018. The experimental results highlight the superior performance and robustness of our proposed model compared to the state-of-the-art approaches. Ablation studies further prove the outstanding performance of all the proposed components. The proposed method provides a promising solution for surgical scene understanding, and opens up a primary step in the Artificial Intelligence (AI)-based VQLA system for surgical training. Our code is available at \href{https://github.com/longbai1006/CAT-ViL}{github.com/longbai1006/CAT-ViL}.

\end{abstract}

\section{Introduction}
\label{sec:1}

Specific knowledge in the medical domain needs to be acquired through extensive study and training. When faced with a surgical scenario, patients, medical students, and junior doctors usually come up with various questions that need to be answered by surgical experts, and therefore, to better understand complex surgical scenarios. However, the number of expert surgeons is always insufficient, and they are often overwhelmed by academic and clinical workloads. Therefore, it is difficult for experts to find the time to help students individually~\cite{sharma2021medfusenet,seenivasan2022surgical}. Automated solutions have been proposed to help students learn surgical knowledge, skills, and procedures, such as pre-recorded videos, surgical simulation and training systems~\cite{hsieh2017vr, lin2006towards}, etc. Although students may learn knowledge and skills from these materials and practices, their questions still need to be answered by experts. Recently, several approaches~\cite{sharma2021medfusenet, seenivasan2022surgical} have demonstrated the feasibility of developing safe and reliable VQA models in the medical field. Specifically, Surgical-VQA~\cite{seenivasan2022surgical} made effective answers regarding tools and organs in robotic surgery, but they were still unable to help students make sense of complex surgical scenarios. For example, suppose a student asks a question about the tool-tissue interaction for a specific surgical tool, the VQA model can only simply answer the question, but cannot directly indicate the location of the tool and tissue in the surgical scene. Students will still need help understanding this complex surgical scene. Another problem with Surgical-VQA is that their sentence-based VQA model requires datasets with annotation in the medical domain, and manual annotation is time-consuming and laborious.

Currently, extensive research and progress have been made on VQA tasks in the computer vision domain~\cite{li2019visualbert}. Models using long-short term memory modules~\cite{yu2019deep}, attention modules~\cite{sharma2021medfusenet}, and Transformer~\cite{li2019visualbert} significantly boost the performance in VQA tasks. Furthermore, FindIt~\cite{kuo2022findit} proposed a unified Transformer model for joint object detection and ViL tasks. However, firstly, most of these models acquire the visual features of key targets through object detection models. In this case, the VQA performance strongly depends on the object detection results, which hinders the global understanding of the surgical scene~\cite{seenivasan2022global}, and makes the overall solution not fully end-to-end. Second, many VQA models employ simple additive, averaging, scalar product, or attention mechanisms when fusing heterogeneous visual and textual features. Nevertheless, in heterogeneous feature fusion, each feature represents different meanings, and simple techniques cannot achieve the best intermediate representation from heterogeneous features. Finally, the VQA model cannot highlight specific regions in the image relevant to the question and answer. Supposing the location of the object in the surgical scene can be known along with the answer by VQLA models, students can compare it with the surrounding tissues, different surgical scenes, preoperative scan data, etc., to better understand the surgical scene~\cite{bai2023surgical}.

In this case, we propose CAT-ViL DeiT for VQLA tasks in surgical scene understanding. Specifically, our contributions are three-fold: (1) We carefully design a Transformer-based VQLA model that can relate the surgical VQA and localization tasks at an instance level, demonstrating the potential of AI-based VQLA system in surgical training and surgical scene understanding. (2) In our proposed CAT-ViL embedding, the co-attention module allows the text embeddings to have instructive interaction with visual embeddings, and the gated module works to explore the best intermediate representation for heterogeneous embeddings. (3) With extensive experiments, we demonstrate the extraordinary performance and robustness of our CAT-ViL DeiT in localizing and answering questions in surgical scenarios. We compare the performance of detection-based and detection-free feature extractors. We remove the computationally costly and error-prone detection proposals to achieve superior representation learning and end-to-end real-time applications. 

\section{Methodology}

\label{sec:2}

\subsection{Preliminaries}
\label{sec:2.1}

\textbf{VisualBERT}~\cite{li2019visualbert} generates text embeddings (including token embedding $e_t$, segment embedding $e_s$, and position embedding $e_p$) based on the strategy of natural language model BERT~\cite{devlin2018bert}, and uses object detection model to extract visual embeddings (consisting of visual features representation $f_v$, segment embedding $f_s$ and position embedding $f_p$). Then, it concatenates visual and text embeddings before feeding the subsequent multilayer Transformer module.

\textbf{Multi-Head Attention}~\cite{vaswani2017attention} can focus limited attention on key and high-value information. In each head $\mathbf{h}_i$, give the certain query $q \in \mathbb{R}^{d_q}$, key matrix $K \in \mathbb{R}^{d_k}$, value matrix $V \in \mathbb{R}^{d_v}$, the attention for each head is calculated as $ \mathbf{h}_i=A\left(\mathbf{W}_i^{(q)} \mathbf{q}, \mathbf{W}_i^{(K)} \mathbf{K}, \mathbf{W}_i^{(V)} \mathbf{V}\right)$. $\mathbf{W}_i^{(q)} \in \mathbb{R}^{p_q \times d_q}$, $\mathbf{W}_i^{(k)} \in \mathbb{R}^{p_k \times d_k}$, $\mathbf{W}_i^{(v)} \in \mathbb{R}^{p_v \times d_v}$ are learnable parameters, and $A$ represents the function of single-head attention aggregation. A linear conversion is then applied for the attention aggregation from multiple heads:
$\mathbf{h}=MA(\mathbf{W}_o\left[
\mathbf{h}_1 \|
\dots \|
\mathbf{h}_h
\right])$.
$\mathbf{W}_o \in \mathbb{R}^{p_o \times h p_v}$ is the learnable parameters in multiple heads. Each head may focus on a different part of the input to achieve the optimal output.

\subsection{CAT-ViL DeiT}
\label{sec:2.2}
\begin{figure*}[]
    \centering
    \includegraphics[width=0.96\linewidth, trim=0 110 410 50]{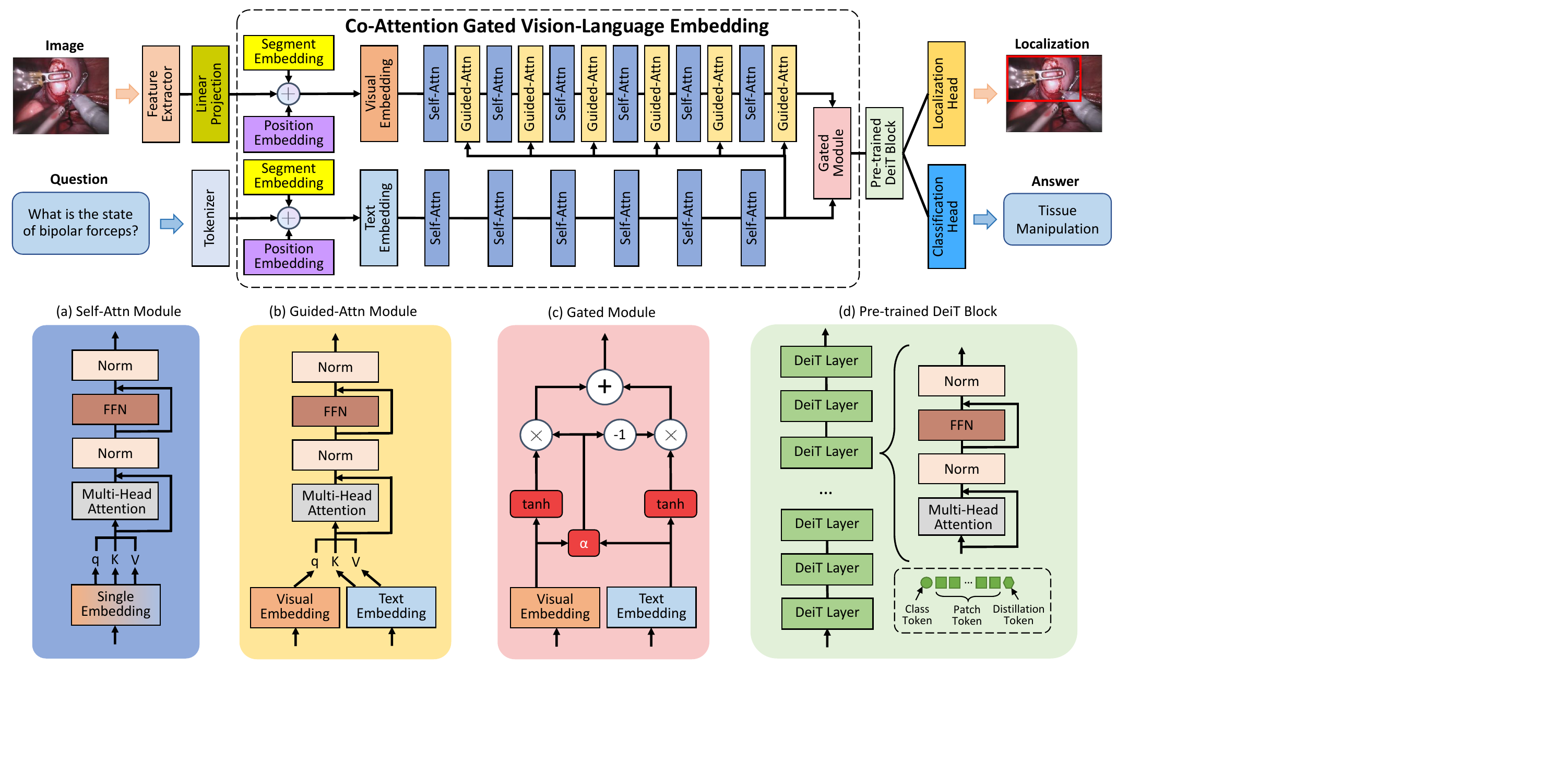}
    \caption{The proposed network architecture. The network components include a visual feature extractor, tokenizer, CAT-ViL embedding module (embedding setup, co-attention learning, gated module), per-trained DeiT block, and task-specific heads. `Attn' represents `Attention'.}
    \label{fig:overview}
\end{figure*}

We present CAT-ViL DeiT to process the information from different modalities and implement the VQLA task in the surgical scene. DeiT~\cite{touvron2021training} serves as the backbone of our network. As shown in Fig.~\ref{fig:overview}, the network consists of a vision feature extractor, a customized trained tokenizer, a co-attention gated embedding module, a standard DeiT module, and task-specific heads. 
\\
\\
\textbf{Feature Extraction:}
\label{sec:2.2.1}
Taking a given image and the associated question, conventional VQA models usually extract visual features via object proposals~\cite{li2019visualbert,yu2019deep}. Instead, we employ ResNet18~\cite{he2016resnet} pre-trained on ImageNet~\cite{deng2009imagenet} as our visual feature extractor. This design enables faster inference speed and global understanding of given surgical scenes. The text embeddings are acquired via a customized pre-trained tokenizer~\cite{seenivasan2022surgical}. The CAT-ViL embedding module then processes and fuses the input embeddings from different modalities.
\\
\\
\textbf{CAT-ViL Embedding:}
\label{sec:2.2.2}
In the following, the extracted features are processed into visual and text embeddings following VisualBERT~\cite{li2019visualbert} as described in Section~\ref{sec:2.1}. However, VisualBERT~\cite{li2019visualbert} and VisualBERT ResMLP~\cite{seenivasan2022surgical} naively concatenate the embeddings from different modalities without optimizing the intermediate representation between heterologous embeddings. In this case, information and statistical representations from different modalities cannot interact perfectly and serve subsequent tasks.

Inspired by~\cite{arevalo2017gmu, yu2019deep}, we replace the naive concatenation operation with a co-attention gated ViL module. The gated module can explore the best combination of the two modalities. Co-attention learning enables active information interaction between visual and text embeddings. Specifically, the guided-attention module is applied to infer the correlation between the visual and text embeddings. The normal self-attention module contains the multi-head attention layer, a feed-forward layer, and ReLU activation. The guide-attention module also contains the above components, but its input is from both two modalities, in which the $q$ is from visual embeddings and $K$,$V$ are from text embeddings:
\begin{equation}
    \mathbf{h}_i=\rm{A}\left(\mathbf{W}_i^{(q)} \mathbf{q}_{visual}, \mathbf{W}_i^{(K)} \mathbf{K}_{text}, \mathbf{W}_i^{(V)} \mathbf{V}_{text}\right)
\end{equation}
Therefore, the visual embeddings shall be reconstructed with the original query, and the key and value of the text embeddings, which can realize the text embeddings to have instructive information interaction with the visual embeddings, and help the model to focus on the targeted image context related to the question. Six guided-attention layers are applied in our network. Thus, the correlation between questions and image regions can be gradually constructed. Besides, we also build six self-attention blocks for both visual and text embeddings to boost the internal relationship within each modality. This step can also avoid `over' guidance and seek a trade-off. Then, the attended text embeddings and text-guided attended visual embedding shall be output from the co-attention module and propagated through the gated module.

Compared to the naive concatenation~\cite{che2022learning}, summation~\cite{wu2022two}, or the multilayer perceptron (MLP) layer~\cite{yu2019deep}, this learnable gated neuron-based model can control the contribution of multimodal input to output through selective activation (set as $tanh$ here). The gate node $\alpha$ is employed to control the weight for selective visual and text embedding aggregation. The equations of the gated module are:
\begin{equation}
    \begin{aligned}  
        \mathbf{E_o} &=\mathbf{w} * \tanh \left(\theta_v\cdot \mathbf{E}_v\right) +(1-\mathbf{w}) * \tanh \left(\theta_t \cdot \mathbf{E}_t\right) \\ 
        \mathbf{w} &=\alpha\left(\theta_\mathbf{w} \cdot\left[\mathbf{E}_v\| \mathbf{E}_t\right]\right) \\ 
    \end{aligned}
    \label{equ:1}
\end{equation}
$\mathbf{E}_v$ and $\mathbf{E}_t$ denotes visual and text embeddings, respectively.
$(\theta_\omega, \theta_v, \theta_t )$ are set as learnable parameters. $[\cdot\,\|\,\cdot]$ means the concatenation operation.  $\mathbf{E_o}$ is the final output embeddings. The activation function internally encodes the text and visual embeddings separately, and the gate weights are used for embedding fusion. This method is uncomplicated and effective, and can optimize the intermediate aggregation of visual and text embeddings while constraining the model.

Subsequently, the fused embeddings $\mathbf{E}_{o}$ shall feed the pre-trained DeiT-Base~\cite{touvron2021training} module before the task-specific heads. The pre-trained DeiT-Base module can learn the joint representation, resolve ambiguous groundings from multimodel information, and maximize performance. 
\\
\\
\textbf{Prediction Heads:}
\label{sec:2.2.3}
The classification head, following the normal classification strategy, is a linear layer with Softmax activation. Regarding the localization head, we follow the setup in Detection with Transformers (DETR)~\cite{carion2020end}. A simple feed-forward network (FFN) with a 3-layer perceptron, ReLU activation, and a linear projection layer is employed to fit the coordinates of the bounding boxes.
The entire network is therefore built end-to-end without multi-stage training.
\\
\\
\textbf{Loss Function:}
\label{sec:2.2.4}
Normally, the cross-entropy loss $\mathcal{L}_{CE}$ serves as our classification loss. The combination of $\mathcal{L}_1$-norm and Generalized Intersection over Union (GIoU) loss~\cite{rezatofighi2019generalized} is adopted to conduct bounding box regression. GIoU loss~\cite{rezatofighi2019generalized} further emphasizes both overlapping and non-overlapping regions of bounding boxes. Then, the final loss function is $\mathcal{L} = \mathcal{L}_{CE} + \left(\mathcal{L}_{GIoU} +\mathcal{L}_{1}\right)$.

\section{Experiments}
\label{sec:3}

\subsection{Dataset}
\label{sec:3.1}
\textbf{EndoVis 2018 Dataset} is a public dataset with 14 robotic surgery videos from MICCAI Endoscopic Vision Challenge~\cite{allan2020endovis18}. The VQLA annotations are publicly accessible by~\cite{bai2023surgical}, in which the QA pairs are from~\cite{seenivasan2022global} and the bounding box annotations are from~\cite{islam2020learning}. Specifically, the QA pairs include 18 different single-word answers regarding organs, surgical tools, and tool-organ interactions. When the question is about organ-tool interactions, the bounding box will contain both the organ and the tool. We follow~\cite{seenivasan2022surgical} to use video [1, 5, 16] as the test set and the remaining as the training set. Statistically, the training set includes 1560 frames and 9014 QA pairs, and the test set has 447 frames and 2769 QA pairs.

\textbf{EndoVis 2017 Dataset} is also a publicly available dataset from the MICCAI Endoscopic Vision Challenge 2017~\cite{allan2019endovis17}, and the annotations are also available by~\cite{bai2023surgical}. We employ this dataset as an external validation dataset to demonstrate the generalization capability of our model in various surgical domains. Specifically, we manually select and annotate frames with common organs, tools, and interactions in EndoVis 2017 Dataset, generating 97 frames with 472 QA pairs. We conduct \textit{no training} but only \textit{testing} on this external validation dataset.

\subsection{Implementation Details}
\label{sec:3.2}
We conduct our comparison experiments against VisualBERT~\cite{li2019visualbert}, VisualBERT ResMLP~\cite{seenivasan2022surgical}, MCAN~\cite{yu2019deep}, VQA-DeiT~\cite{touvron2021training}, MUTAN~\cite{ben2017mutan}, MFH~\cite{yu2018beyond}, and BlockTucker~\cite{ben2019block}. In VQA-DeiT, we use pre-trained DeiT-Base block~\cite{touvron2021training} to replace the multilayer Transformer module in VisualBERT~\cite{li2019visualbert}. To keep a fair comparison of VQLA tasks, we use the same prediction heads in and loss function in Section~\ref{sec:2.2}. The evaluation metrics are accuracy, f-score, and mean intersection over union (mIoU)~\cite{rezatofighi2019generalized}. All models are trained on NVIDIA RTX 3090 GPUs using Adam optimizer~\cite{kingma2014adam} with PyTorch. The epoch, batch size, and learning rate are set to $80$, $64$, and $1 \times 10^{-5}$, respectively. The experimental results are the average results with five different random seeds.

\subsection{Results}
\label{sec:3.3}

\begin{figure*}[t]
    \centering
    \includegraphics[width=.92\linewidth, trim=0 430 500 0]{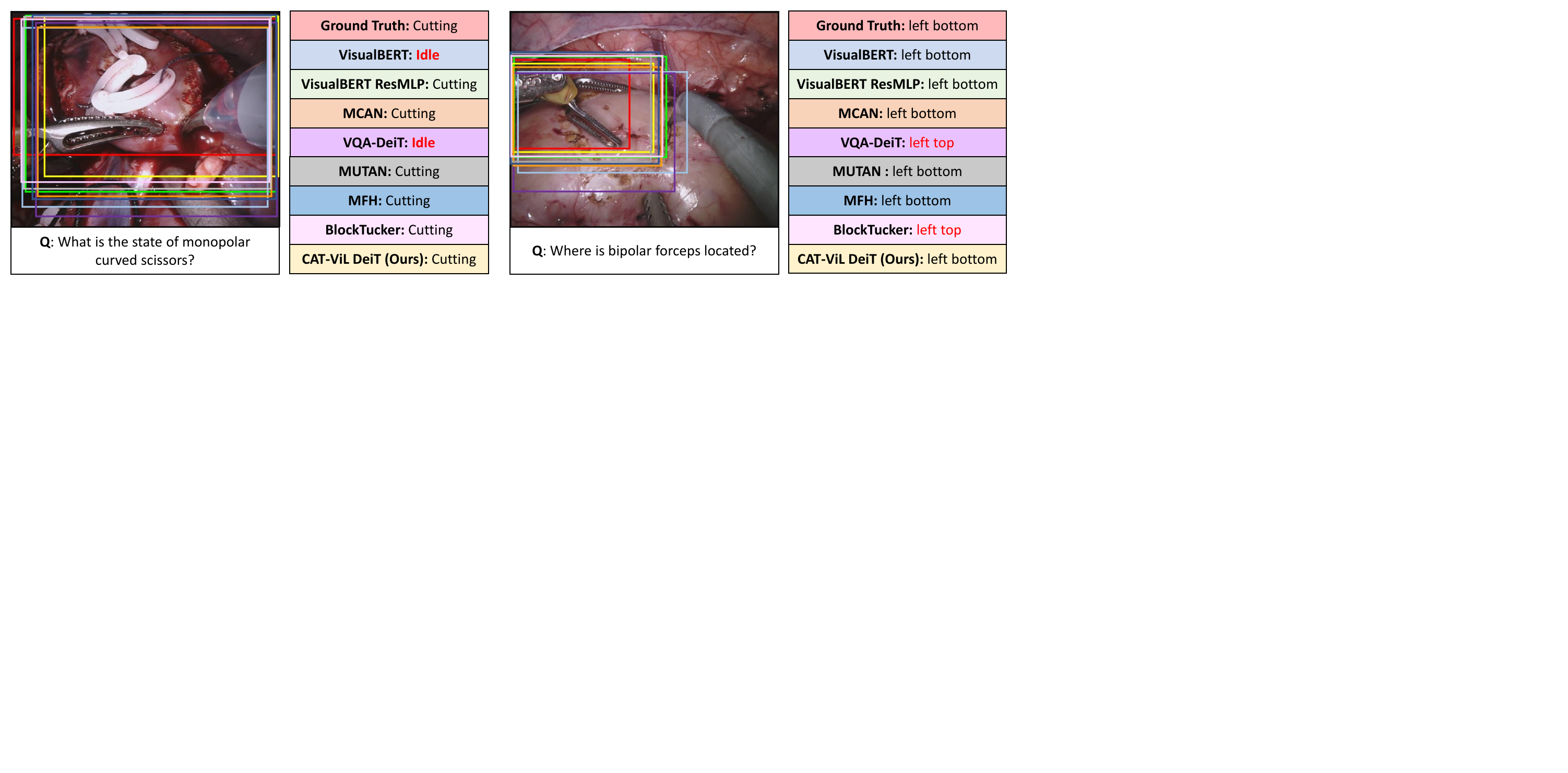}
    \caption{
    Qualitative comparison on the VQLA task. Our CAT-ViL DeiT (Yellow) displays state-of-the-art (SOTA) performance on generating the answers and location against VisualBERT (light blue)~\cite{li2019visualbert}, VisualBERT ResMLP (green)~\cite{seenivasan2022surgical}, MCAN (orange)~\cite{yu2019deep}, VQA-DeiT (purple)~\cite{touvron2021training}, MUTAN (gray)~\cite{ben2017mutan}, MFH (dark blue)~\cite{yu2018beyond}, and BlockTucker (pink)~\cite{ben2019block}. The Ground Truth bounding box is red.
    }
    \label{fig:visualization}
\end{figure*}

\begin{table}[t]
\caption{Comparison experiments on EndoVis-18 and EndoVis-17 datasets.}
\centering
\resizebox{\textwidth}{!}{
\begin{tabular}{c|c|c|ccc|ccc}
\noalign{\smallskip}\hline
\multirow{2}{*}{Models} & \multicolumn{2}{c|}{Visual Feature} & \multicolumn{3}{c|}{EndoVis-18} & \multicolumn{3}{c}{EndoVis-17} \\ \cline{2-9}
& Detection & \multicolumn{1}{c|}{\makecell[c]{Inference \\ Speed}} & Acc & F-Score & mIoU & Acc & F-Score & mIoU \\ \hline
VisualBERT~\cite{li2019visualbert} & \multirow{8}{*}{FRCNN~\cite{ren2015faster}} & \multirow{8}{*}{55.28 ms} & 0.5973 & 0.3223 & 0.7340 & 0.4382 & \textbf{0.3743} & 0.6822 \\
VisualBERT R~\cite{seenivasan2022surgical} & & & 0.6064 & 0.3226 & 0.7305 & 0.4267 & 0.3506 & 0.6947               \\
MCAN~\cite{yu2019deep} & & & 0.6084 & 0.3428 & 0.7257 & 0.4258 & 0.3035 & 0.6832               \\
VQA-DeiT~\cite{touvron2021training} & & & 0.6089 & 0.3217 & 0.7338 & 0.4492 & 0.3213 & 0.7134               \\
MUTAN~\cite{ben2017mutan} & & & 0.6049 & 0.3238 & 0.7217 & 0.4364 & 0.3206 & 0.6870 \\
MFH~\cite{yu2018beyond} & & & 0.6179 & 0.3158 & 0.7227 & 0.3729 & 0.2048 & \textbf{0.7183} \\
BlockTucker~\cite{ben2019block} & & & 0.6067 & 0.3414 & 0.7313 & 0.4364 & 0.3210 & 0.6825 \\
CAT-ViL DeiT \textbf{(Ours)} & & & \textbf{0.6192} & \textbf{0.3521} & \textbf{0.7482} & \textbf{0.4555} & 0.3676 & 0.7049           \\ \cline{1-9} 
VisualBERT~\cite{li2019visualbert}& \multirow{8}{*}{\XSolidBrush} & \multirow{8}{*}{6.64 ms} & 0.6268 & 0.3329 & 0.7391 & 0.4005 & 0.3381 & 0.7073  \\
VisualBERT R~\cite{seenivasan2022surgical} & & & 0.6301 & 0.3390 & 0.7352 & 0.4190 & 0.3370 & 0.7137 \\
MCAN~\cite{yu2019deep} & & & 0.6285 & 0.3338 & 0.7526 & 0.4137 & 0.2932 & 0.7029 \\
VQA-DeiT~\cite{touvron2021training} & & & 0.6104 & 0.3156 & 0.7341 & 0.3797 & 0.2858 & 0.6909 \\
MUTAN~\cite{ben2017mutan} & & & 0.6283 & \textbf{0.3395} & 0.7639 & 0.4242 & 0.3482 & 0.7218 \\
MFH~\cite{yu2018beyond} & & & 0.6283 & 0.3254 & 0.7592 & 0.4103 & 0.3500 & 0.7216 \\
BlockTucker~\cite{ben2019block} & & & 0.6201 & 0.3286 & 0.7653 & 0.4221 & 0.3515 & 0.7288 \\
CAT-ViL DeiT \textbf{(Ours)} & & & \textbf{0.6452} & 0.3321 & \textbf{0.7705} & \textbf{0.4491} & \textbf{0.3622} & \textbf{0.7322} \\ \hline

\end{tabular}}
\label{tab:main}
\end{table}

\begin{table}[t]
\caption{Ablation study on different fusion strategies. All experiments use the same feature extractor, DeiT backbone, and prediction heads. `Attn' denotes `Attention'.}
\centering
\resizebox{0.97\textwidth}{!}{
\begin{tabular}{lcccccc}
\hline\noalign{\smallskip} 
\multirow{2}{*}{Fusion Strategies} & \multicolumn{3}{c}{EndoVis-18}        & \multicolumn{3}{c}{EndoVis-17}        \\
& Acc & F-Score & mIoU & Acc & F-Score & mIoU              \\ \hline\noalign{\smallskip}
Concatenation~\cite{li2019visualbert} & 0.6104 & 0.3156 & 0.7341 & 0.3797 & 0.2858 & 0.6909   \\
JCA~\cite{praveen2022joint} & 0.6024 & 0.3010 & 0.7527 & 0.3750 & 0.2835 & 0.7145 \\
MMHCA~\cite{georgescu2023multimodal} & 0.6096 & 0.3124 & 0.7449 & 0.3581 & 0.3001 & 0.7077 \\
MAT~\cite{wu2022multimodal} & 0.6186 & 0.3179 & 0.7415 & 0.3369 & 0.2850 & 0.6956 \\
Gated Fusion~\cite{arevalo2017gmu} & 0.6071 & \textbf{0.3793} & 0.7683 & 0.4030 & 0.2824 & \textbf{0.7388} \\
Self-Attn~\cite{vaswani2017attention} & 0.5923 & 0.3095 & 0.7271 & 0.3686 & 0.2673 & 0.6718          \\
Guided-Attn~\cite{yu2019deep} & 0.6194 & 0.3134 & 0.7310 & 0.3517 & 0.2290 & 0.7185          \\
Co-Attn (Bi) & 0.6056 & 0.3090 & 0.7206 & 0.3644 & 0.3083 & 0.7044 \\
Co-Attn (V2T) & 0.6392 & 0.3263 & 0.7218 & 0.3453 & 0.2265 & 0.7143 \\
Co-Attn (T2V)~\cite{yu2019deep} & 0.6136 & 0.3208 & 0.7273 & 0.3805 & 0.3026 & 0.6870          \\
Self-Attn Gated \textbf{(Ours)} & 0.6249 & 0.3078 & 0.7314 & 0.3263 & 0.2897 & 0.7086          \\
Guided-Attn Gated \textbf{(Ours)} & 0.6280 & 0.3127 & 0.7651 & 0.3962 & 0.3337 & 0.7145  \\
CAT-ViL (Bi) \textbf{(Ours)} & 0.6230 & 0.3121 & 0.7415 & 0.4258 & 0.3593 & 0.7282 \\
CAT-ViL (V2T) \textbf{(Ours)} & 0.6352 & 0.3259 & 0.7600 & 0.4301 & 0.3543 & 0.7074 \\
CAT-ViL (T2V) \textbf{(Ours)} & \textbf{0.6452} & 0.3321 & \textbf{0.7705} & \textbf{0.4491} & \textbf{0.3622} & 0.7322 \\ \noalign{\smallskip}
\hline
\label{table:ablation}
\end{tabular}}
\end{table}

\begin{figure*}[]
    \centering
    \includegraphics[width=\linewidth]{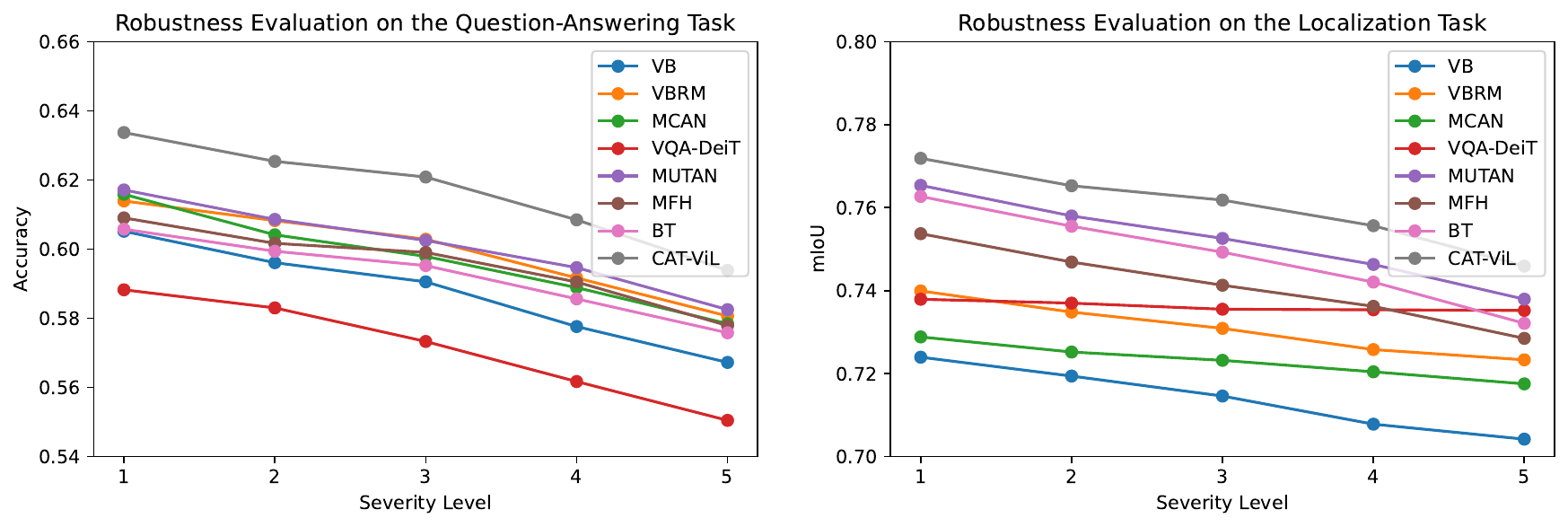}
    \caption{Robustness experiments on the EndoVis-18 dataset. We process the data with 18 corruption methods at each severity level and average the prediction results.
    }
    \label{fig:robustness}
\end{figure*}

Fig.~\ref{fig:visualization} presents the visualization and qualitative comparison of the surgical VQLA system. Quantitative evaluation in Table~\ref{tab:main} presents that our proposed model using ResNet18~\cite{he2016resnet} feature extractor suppresses all SOTA models significantly. Additionally, we compare the performance between using object proposals (Faster RCNN~\cite{ren2015faster}) and using features from the entire image (ResNet18~\cite{he2016resnet}). The experimental results in EndoVis-18 show that removing the object proposal model improves the performance appreciably on both question-answering and localization tasks, which demonstrates the impact of this approach in correcting potential false detections. Meanwhile, in the external validation set -  EndoVis-17, our CAT-ViL DeiT with RCNN feature extractor suffers from domain shift and class imbalance problems, thus achieving poor performance. However, our final model, CAT-ViL DeiT with ResNet18 feature extractor, endows the network with global awareness and outperforms all baselines in terms of accuracy and mIoU, proving the superiority of our method. The inference speed is also enormously accelerated, demonstrating its potential in real-time applications.   

Furthermore, a robustness experiment is conducted to observe the model stability when test data is corrupted. We set 18 types of corruption on the test data based on the severity level from 1 to 5 by following~\cite{hendrycks2019benchmarking}. Then, the performance of our model and all comparison methods on each corruption severity level is presented in Fig.~\ref{fig:robustness}. As the severity increases, the performance of all models degrades. However, our model shows good stability against corruption, and presents the best prediction results at each severity level. The excellent robustness of our model brings great potential for real-world applications.

Finally, we conduct an ablation study on different ViL embedding techniques with the same feature extractors and DeiT backbone in Table~\ref{table:ablation}. We compare with Concatenation~\cite{li2019visualbert}, Joint Cross-Attention (JCA)~\cite{praveen2022joint}, Multimodal Multi-Head Convolutional Attention (MMHCA)~\cite{georgescu2023multimodal}, Multimodal Attention Transformers (MAT)~\cite{wu2022multimodal}, Gated Fusion~\cite{arevalo2017gmu}, Self-Attention Fusion~\cite{vaswani2017attention}, Guided-Attention Fusion~\cite{yu2019deep}, Co-Attention Fusion (T2V: Text-Guide-Vision)~\cite{yu2019deep}. Besides, we explore the Co-Attention module with different directions (V2T: Vision-Guide-Text, and Bidirectional). Furthermore, we also incorporate the Gated Fusion with different attention mechanisms (Self-Attention, Guided-Attention, Bidirectional Co-Attention, Co-Attention (V2T), Co-Attention (T2V)) for detailed comparison. They are shown as `Self-Attn Gated', `Guided-Attn Gated',  `CAT-ViL (Bi)', `CAT-ViL (V2T)' and `CAT-ViL (T2V)' in Table~\ref{table:ablation}. The study proves the superior performance of our ViL embedding strategy against other advanced methods. We also demonstrate that integrating attention feature fusion techniques and the gated module will bring performance improvement.

\section{Conclusions}
\label{sec:4}

This paper presents a Transformer model with CAT-ViL embedding for the surgical VQLA tasks, which can give the localized answer based on a specific surgical scene and associated question. It brings up a primary step in the study of VQLA systems for surgical training and scene understanding. The proposed CAT-ViL embedding module is proven capable of optimally facilitating the interaction and fusion of multimodal features. Numerous comparative, robustness, and ablation experiments display the leading performance and stability of our proposed model against all SOTA methods in both question-answering and localization tasks, as well as the potential of real-time and real-world applications. Furthermore, our study opens up more potential VQA-related problems in the medical community. Future work can be focused on quantifying and improving the reliability and uncertainty of these safety-critical tasks in the medical domain.

\subsubsection{Acknowledgements.}

This work was funded by Hong Kong RGC CRF C4063-18G,  CRF C4026-21GF, RIF R4020-22, GRF 14203323, GRF 14216022, GRF 14211420, NSFC/RGC JRS N\_CUHK420/22; Shenzhen-Hong Kong-Macau Technology Research Programme (Type C 202108233000303); Guangdong GBABF \#2021B1515120035. M. Islam was funded by EPSRC grant [EP/W00805X/1].

\bibliography{reference}{}
\bibliographystyle{splncs04}                                                                                                                                                                                   
                                       
\newpage
\section*{Supplementary Materials for ``CAT-ViL: Co-Attention Gated Vision-Language Embedding for Visual Question Localized-Answering in Robotic Surgery''}

\begin{figure}
\centering
\includegraphics[width=\textwidth, trim=10 610 370 0]{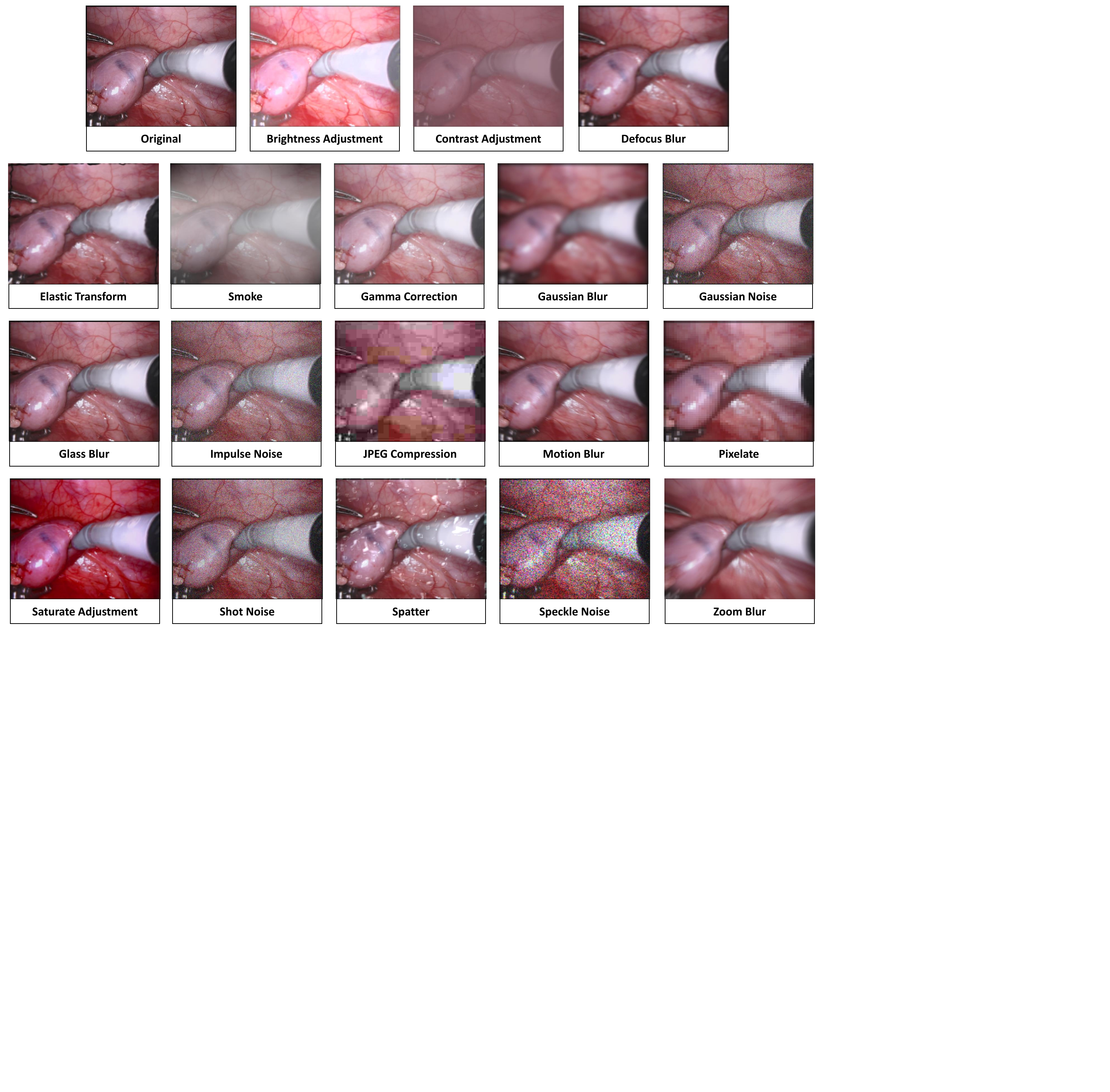}
\caption{Visualization of our corrupted data for robustness experiment.} \label{figa1}
\end{figure}

\begin{table}[!h]
\caption{Ablation Study on different co-attention layers.}
\centering
\resizebox{0.87\textwidth}{!}{
\begin{tabular}{ccccccc}
\hline\noalign{\smallskip} 
\multirow{2}{*}{Number of Layers} & \multicolumn{3}{c}{EndoVis-18}        & \multicolumn{3}{c}{EndoVis-17}        \\
& Acc & F-Score & mIoU & Acc & F-Score & mIoU              \\ \hline\noalign{\smallskip}
2 & 0.6212 & 0.3100 & 0.7686 & 0.4573 & 0.3399 & \textbf{0.7352} \\
4 & 0.6255 & \textbf{0.3346} & 0.7550 & 0.4364 & 0.3402 & 0.7176 \\
6 \textbf{(Ours)} & \textbf{0.6452} & 0.3321 & \textbf{0.7705} & 0.4491 & \textbf{0.3622} & 0.7322 \\
8 & 0.6355 & 0.3070 & 0.7696 & \textbf{0.4619} & 0.3265 & 0.7246 \\
10 & 0.6306 & 0.3135 & 0.7696 & 0.3877 & 0.3023 & 0.7258 \\ \noalign{\smallskip}
\hline
\label{table:ablation2}
\end{tabular}}
\end{table}



\end{document}